\documentclass[runningheads]{llncs}
\usepackage{amsmath}
\usepackage{graphicx}
\usepackage{algorithm}
\usepackage{algorithmic}
\usepackage{multirow}
\usepackage{subcaption}
\usepackage{booktabs}

\usepackage[width=122mm,left=12mm,paperwidth=146mm,height=193mm,top=12mm,paperheight=217mm]{geometry}

\begin{document}
\pagestyle{headings}
\mainmatter
\def\ECCV14SubNumber{111}  
\title{Stereo Matching by Joint Energy Minimization}

\author{Hongyang Xue \and Deng Cai}

\institute{State Key Lab of CAD\&CG \\ College of Computer Science \\ Zhejiang University, China \\  \email{dengcai@cad.zju.edu.cn} }

\date{}

\maketitle

\begin{abstract}
	In \cite{mozerov2015accurate}, Mozerov et al. propose to perform stereo matching as a two-step energy minimization problem. For the first step they solve a fully connected MRF model. And in the next step the marginal output is employed as the unary cost for a locally connected MRF model.
	
	In this paper we intend to combine the two steps of energy minimization in order to improve stereo matching results. We observe that the fully connected MRF leads to smoother disparity maps, while the locally connected MRF achieves superior results in fine-structured regions. Thus we propose to jointly solve the fully connected and locally connected models, taking both their advantages into account. The joint model is solved by mean field approximations. While remaining efficient, our joint model outperforms the two-step energy minimization approach in both time and estimation error on the Middlebury stereo benchmark v3. 
	
\end{abstract}

\section{Introduction}
Stereo matching is one of the fundamental approaches to acquire depth information \cite{brown2003advances}, \cite{nair2013survey}, \cite{scharstein2002taxonomy}. It is important in many fields such as autonomous driving \cite{geiger2012we}, \cite{koo2015robust}, robotics \cite{nalpantidis2007review}, \cite{song2013survey} and 3D scene reconstruction \cite{bleyer2012extracting}, \cite{kim20133d}, \cite{vzbontar2015stereo}.  Given a pair of images, the task of stereo matching is to assign a disparity value to each pixel of the scene. Thus stereo matching is actually a discrete labeling problem \cite{scharstein2002taxonomy}, \cite{yamaguchi2014efficient} where we need to find a solution which labels every pixel with a disparity value and satisfies:
\begin{itemize}
	\item label costs are minimized
	\item disparities are spatially smooth
\end{itemize}

Mozerov et al. \cite{mozerov2015accurate} proposed a two-step energy minimization algorithm where two random fields are solved sequentially to obtain the disparity map. In their work, label costs are encoded in unary potentials and spatially smooth solution is enforced by pairwise potentials. They solve a fully connected MRF model with cost filtering approaches. Then they take the marginal function as the unary cost for the next locally connected MRF model. 

The consecutive applications of the two MRF models are mainly proposed to improve the stereo results in occluded regions \cite{mozerov2015accurate}. Besides occlusions, we observe that the fully connected MRF based(FCM) energy minimization usually obtains good stereo results in large, flat and simple structures of the scene because it takes the global information into consideration. They usually lead to smoother disparity maps. However, they tend to fail in occluded regions and cause errors in fine-structured areas where locally connected MRF(LCM) based models achieve better results. LCM based energy minimization preserves more fine structures because they only care about neighboring pixels and thus avoid over-smoothing. But LCM usually leads to noisy and unsmooth disparity maps because it is somewhat \textit{short-sighted} (see Figure 1). This effect is similar to the results of different matching window sizes in window-based matching approaches \cite{kanade1994stereo}. Though consecutive applications of the two models can improve the quality of the disparity map, the two-step model cannot inhibit their faults simultaneously. It is apparent and reasonable to consider both models at the same time.

\begin{figure}[!htbp]
		\begin{minipage}[t]{0.33\linewidth}
			\centering
			\includegraphics[width=1.5in]{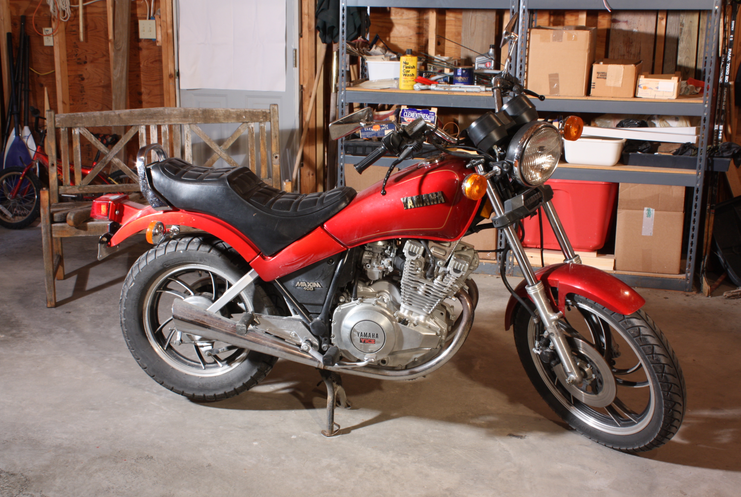}
			\label{fig:side:a111}
			\subcaption{Left}
		\end{minipage}%
	\begin{minipage}[t]{0.33\linewidth}
		\centering
		\includegraphics[width=1.5in]{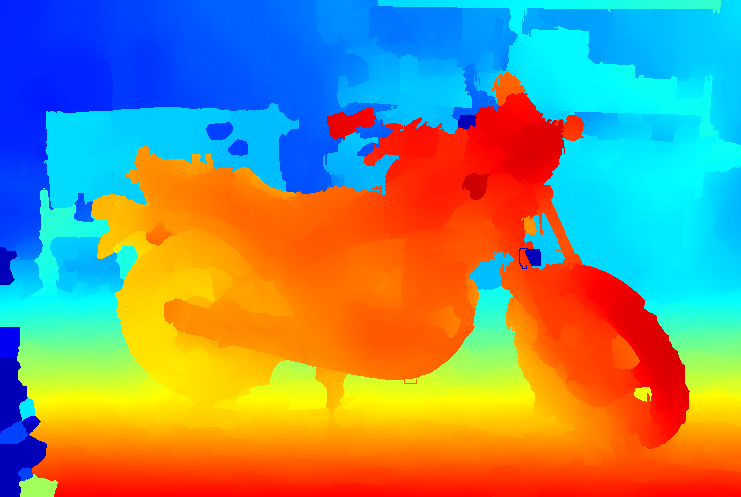}
		\label{fig:side:d111}
		\subcaption{FCM}
	\end{minipage}%
	\begin{minipage}[t]{0.33\linewidth}
		\centering
		\includegraphics[width=1.5in]{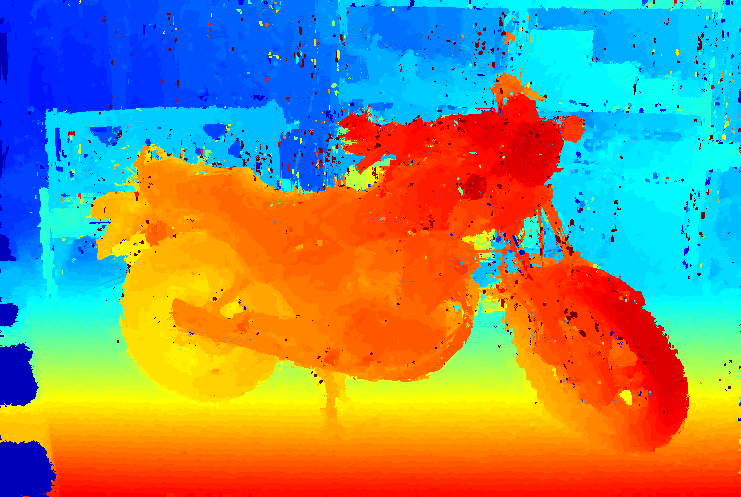}
		\label{fig:side:c111}
		\subcaption{LCM}
	\end{minipage}
	\caption{FCM achieves smoother disparity maps than LCM. But LCM preserves more fine structures.}
\end{figure}
 
  The inference of markov random fields is not a trivial task. A significant amount of research has been focused on inferencing markov random fields \cite{Sutton2010An}, \cite{Wang2013Markov}. Exact inference of fully connected MRF models is usually considered intractable due to unsolvable computational complexity \cite{mozerov2015accurate}, \cite{rhemann2011fast}. Therefore, conventional stereo methods based on energy minimization are always defined on locally connected models \cite{yang2009stereo}. A lot of research such as graph cut \cite{boykov2001fast} and loopy belief propagation \cite{sun2003stereo}, have been devoted to tractable approximations. These approximations suffer from a drawback that they are often very slow and usually introduce large errors, making the solutions less-global optimal \cite{rhemann2011fast}. 
 
 Luckily, under the situation of Gaussian edge potentials, an efficient solution to fully connected CRFs was proposed in \cite{krahenbuhl2012efficient} with accelerated message passing by high-dimensional filtering \cite{adams2010fast}. The fast solution to fully connected random fields has been proved to be effective and efficient in semantic image segmentation tasks. Zheng et al. \cite{zheng2015conditional} went further and formulated the mean field algorithm as a CNN-RNN network. In addition, Mozerov et al. \cite{mozerov2015accurate} also employ high-dimensional filtering \cite{adams2010fast} for their cost filtering approach.

Based on the two-step energy minimization approach(TSGO) \cite{mozerov2015accurate}, we propose a joint energy minimization approach(JEM) where we jointly solve the fully connected MRF and the locally connected MRF models. We come up with a random field which contains a fully connected pairwise potential as well as a locally connected one. We employ the mean field approximation to the joint model and develop an algorithm based on \cite{krahenbuhl2012efficient}. To the best of our knowledge, we are the first to employ mean field inference with high-dimensional filtering \cite{krahenbuhl2012efficient} to solve the fully connected MRF models for stereo matching. The experimental results show that our joint model outperforms TSGO \cite{mozerov2015accurate} on the Middlebury benchmark v3.

The remainder of the paper is organized as follows. The joint model will be explained in Section 2. The fully connected part will be explained in Section 2.1, and the locally connected part will be explained in Section 2.2. In Section 3, the mean field inference solution to our joint model will be derived and explained. In section 4 we will discuss post-processing techniques. In section 5 we will display experiments and results.

\section{Problem Definition}

Given a pair of rectified images, we wish to compute the disparity $d$ for each pixel in the left image \cite{vzbontar2015stereo}. Disparity refers to the difference in horizontal location of an object in the left and right image \cite{marr1976cooperative}. Assume there are $M$ disparity levels, for a pair of images, disparity of each pixel takes finite discrete values from $\mathcal{D} = \{0,1,\cdots,M-1\}$.

The stereo matching problem is usually formulated on a random field $\mathbf{D}$ defined over a set of variables $\{D_1,\cdots,D_N\}$ and the domain of each variable is $\mathcal{D}$ \cite{sun2003stereo}. The Gibbs energy of a labeling $\mathbf{d} \in \mathcal{D}^N$ is

\begin{equation}
E(\mathbf{d)}=\sum_{i\in\mathcal{V}}\psi_u(d_i) + \sum_{(i,j)\in\mathcal{E}}\Psi_p(d_i,d_j)
\end{equation}
where the set $\mathcal{V}$ corresponds to pixels and set $(i,j)\in\mathcal{E}$ to edges of an image graph $\mathcal{G}=(\mathcal{E},\mathcal{V})$; $d_i$ denotes the disparity of pixel $i$ and belongs to the discrete set of disparities $\mathcal{D}$. $\psi_u$ is the unary term which takes the conventional penalty cost; $\Psi_p(i,j)$ is the pairwise potential which encodes the interactions between pixels $(i,j)$.

The choice of unary potential is a key in the process of stereo matching \cite{hirschmuller2007evaluation}. There are many kinds of costs, from simple per-pixel matching dissimilarity measure and non-parametric transforms \cite{zabih1994non}, \cite{lewis1995fast} with a support region, to costs calculated by convolutional neural networks \cite{vzbontar2015stereo}. In \cite{mozerov2015accurate}, the authors argue that cost with a support region is redundant for their fully connected model. In their approach, the result of FCM only provides a better unary cost for the LCM, so the unary cost with a support region is redundant for FCM. In our model, both the FCM and LCM utilizes the same unary cost. Therefore in our joint model, we apply the cost by gradient image and Census transform \cite{hirschmuller2002real} which is used in semi-global matching \cite{hirschmuller2005accurate} for simplicity and sufficiency. Notice that it is convenient to employ more sophisticated and representative costs to our model.

In our model, the pairwise potential consists of two parts: the fully connected pairwise $\psi_p(d_i,d_j)$ and the locally connected ones $\tilde{\psi}_p(d_i,d_j)$. We can write the full model as:

\begin{equation}
E(\mathbf{d})=\sum_{i}\psi_u(d_i) + \sum_i\sum_{j>i}\psi_p(d_i,d_j)+ \sum_{i}\sum_{j\in\mathcal{N}(i)}\tilde{\psi}_p(d_i,d_j)
\end{equation}

The problem is to find $\mathbf{d}$ that minimizes the above energy. Following conventional formulation, we transform the energy into Gibbs distribution, where $Z$ is the partition function and $\mathbf{X}$ is the observation of the image pair.

\begin{equation}
P(\mathbf{d}|\mathbf{X})=\frac{1}{Z}\mathrm{e}^{-E(\mathbf{d})}
\end{equation}

In the other word, we need to do MAP(maximum a priori) inference to the Gibbs distribution $P(\mathbf{d}|\mathbf{X})$ given images $\mathbf{X}$. For convenience we will omit $\mathbf{X}$ in the remaining part of this paper.
\begin{equation}
\arg \max_{\mathbf{d}} P(\mathbf{d})
\end{equation}

\subsection{Fully Connected Pairwise Potential}
The fully connected pairwise potential in our joint model is
\begin{equation}
\psi_{p}(d_{i},d_{j})=\omega \cdot \mu(d_i,d_j)\cdot \exp({-\frac{|x_i-x_j|^{2}}{2\sigma_{x}^{2}}}-\frac{|f_i-f_j|^{2}}{2\sigma_f^{2}})
\end{equation}
where $\mu(d_i,d_j)$ is the label compatibility function and we adopt the widely used Potts model \cite{scharstein2007learning}, $x_i$ is the position of pixel $i$ and $f_i$ is the feature of pixel $i$. We take $x_i$ the coordinate of the pixel and $f_i$ the color vector of the pixel.

This term is a Potts model bilateral filter and called Gaussian edge potential in \cite{krahenbuhl2012efficient}. It takes into account the interactions of every pair of pixels in the image. The Potts model implies that when two pixels have the same disparity value, their pairwise potential will not be considered. The bilateral kernel enforces smoothness in both color(appearance) and spatial domain while preserving edges \cite{elad2002origin}, \cite{yang2012non}. The appearance term is inspired from the observation that pixels with similar colors tend to own the same disparity value. The spatial kernel confines the influence of the appearance kernel within some spatial range.

We denote the bilateral filter as $k(i,j)$, that is, 

\begin{equation}
k(i,j) = \exp({-\frac{|x_i-x_j|^{2}}{2\sigma_{x}^{2}}}-\frac{|f_i-f_j|^{2}}{2\sigma_f^{2}})
\end{equation}

In semantic image segmentation \cite{krahenbuhl2012efficient}, the pairwise potential contains a Gaussian kernel and a bilateral kernel. Noting that different from the bilateral kernel, the Gaussian kernel only enforces spatial smoothness which is also addressed in the locally connected potential in our model. To reduce redundancy we omit the fully connected Gaussian pairwise energy.

\subsection{Locally Connected Pairwise Potential}
The locally connected pairwise potential (7) is the same as the one in \cite{mozerov2015accurate}. It only takes into account a small neighborhood of each pixel and penalizes disparity unsmoothness in a small region. It enforces local smoothness of the disparity map:

\begin{equation}
\tilde{\psi}_p=\tilde{\omega}\cdot \tilde{w}(i,j)\cdot \tilde{\phi}(d_i,d_j)
\end{equation} 

following \cite{boykov2001fast} and \cite{mozerov2015accurate}, the function $\tilde{w}(i,j)$ divides the set of edges into two or more subsets corresponding to a high or low image gradient:

\begin{equation}
\tilde{w}(i,j)=\begin{cases}
\lambda_1 \; & |\mathbf{I}_i-\mathbf{I}_j| < \mu_1 \\
\lambda_2 \; & \mu_1\le |\mathbf{I}_i-\mathbf{I}_j| < \mu_2 \\
\lambda_3 \; & |\mathbf{I}_i-\mathbf{I}_j| \ge \mu_2
\end{cases}
\end{equation}

where

$$
|\mathbf{I}_i-\mathbf{I}_j| = \sum_{c\in \{r,g,b\}}|I^c_i-I^c_j|
$$

The weights in (8) implies strong dependency between nodes when the colors are similar. The smoothness multiplier $\tilde{\phi}(d_i,d_j)$ is chosen as:
\begin{equation}
\tilde{\phi}(d_i,d_j)=\begin{cases}
0 \; & |d_i-d_j|=0 \\
\beta \; & |d_i-d_j| = 1 \\
1 \; & |d_i-d_j| > 1
\end{cases}
\end{equation}

As pointed out in \cite{mozerov2015accurate}, a nonzero value of $|d_i-d_j|$ is usually considered as the discontinuity of the disparity map and this is not true in general because a value 1 for $|d_i-d_j|$ could be caused by discretization errors. Thus the case $|d_i-d_j| = 1$ is specifically picked out and assigned a small penalization $\beta$. We also tried to pick out the case $|d_i-d_j| = 2$ and experiments show that adding more pieces to the multiplier function is useless.
\section{Inference}
Our model (2) contains a fully connected energy term which is intractable to find an exact solution \cite{koller2009probabilistic}. In \cite{krahenbuhl2012efficient}, a fully connected CRF is applied for semantic image segmentation where Krahenbuhl et al. employ mean field approximation with high-dimensional filtering for the inference.
Following their idea, we also employ mean field approximation to inference our joint model.

Mean Field Inference with high-dimensional filtering has been proved in \cite{krahenbuhl2012efficient} to be an efficient solver for dense connected CRF models.
The basic idea of Mean Field Approximation is that instead of computing the exact distribution $P(\mathbf{D})$, we approximate the original distribution with a factored distribution $Q(\mathbf{D})$ that minimizes the KL-divergence $\mathbf{D}(Q||P)$ \cite{koller2009probabilistic}. The factored distribution $Q(\mathbf{D})$ is the product of distribution on individual pixels:

\begin{equation}
Q(\mathbf{D})=\prod_iQ_i(D_i)
\end{equation}

When minimizing the KL-divergence, the mean field approximation yields an iterative update equation:

\begin{equation}
\begin{aligned}
Q_i(d_i)\leftarrow \exp\{\psi_{u}(d_i)- & \sum_{j\ne i}\sum_{d_j\in \mathrm{Val}(D_j)}\psi_p(d_i,d_j)Q_j(d_j) \\ & -\sum_{j\in \mathcal{N}_i}\sum_{d_j\in \mathrm{Val}(D_j
)}\tilde{\psi}_p(d_i,d_j)Q_j(d_j)\}
\end{aligned}
\end{equation}

Based on this iterative update equation, we extend the inference algorithm in \cite{krahenbuhl2012efficient} which has meassage passing, compatibility transform and local update steps to solve our joint model(see Algorithm \ref{alg_1}).

\begin{algorithm}[t]
	\caption{Mean Field in our joint model}
	Initialize $Q_{i}(d_i)\leftarrow \frac{1}{Z_i}\exp(-\psi_u(d_i))$ \\
	\textbf{while} not converged \textbf{do}
	\begin{enumerate}
		\item $\tilde{Q}_i(l)\leftarrow \sum_{j\ne i}k(i,j)Q_j(l)$, $\tilde{P}_i(l)\leftarrow \sum_{j\in\mathcal{N}(i)}\tilde{w}(i,j)Q_j(l)$
		\item $\hat{Q}_i(d_i)\leftarrow \sum_{l\in \mathcal{D}}\omega \cdot \mu(d_i,l)\tilde{Q}_i(l)$, $\hat{P}_i(d_i)=\sum_{l\in \mathcal{D}}\tilde{\omega}\cdot \tilde{\phi}(d_i,l)\tilde{P}_i(l)$
		\item $Q_i(d_i)=\exp(-\psi_u(d_i)-\hat{Q}_i(d_i)-\hat{P}_i(d_i))$
		\item normalize $Q_i(d_i)$
	\end{enumerate}
	\textbf{end} \textbf{while}
	\label{alg_1} 
\end{algorithm}

 After the marginal of the factored distribution is computed on each pixel, we employ the winner-take-all \cite{mei2013segment} strategy to decide the disparity of each pixel.
 \begin{equation}
 d_i=\arg \max_d Q_i(d)
 \end{equation}
 \subsection{Time Complexity Analysis}
 One iteration of the algorithm has time complexity $\mathcal{O}(MN)$ where $N$ is number of pixels and $M$ is the total number of disparity levels. It is obvious that the second(Compatibility transform), third(Local update) and fourth step in the iteration of Algorithm \ref{alg_1} has complexity $\mathcal{O}(MN)$. The first step(Message passing) seems to have complexity quadratic in $N$. Luckily in \cite{krahenbuhl2012efficient} Krahenbuhl et al. propose to employ a trick of fast high-dimensional filtering which accelerates the message passing step to $\mathcal{O}(MN)$. 

\section{Post-processing}
Post-processing steps are indispensable for obtaining good stereo results. All sate-of-the-art algorithms in stereo use post-processing techniques. Some post processing approaches are fast, while the others can be quite time-consuming. We utilize several post-processing techniques. First we follow the post processing methods in \cite{mozerov2015accurate} and take only the first step: Left-to-right disparity map cross-checking(LRC). Experimental results show that our approach outperforms the two-step algorithm even with only this single step of post-processing. 

Besides the LRC step, we also employ several other techniques to post-process our stereo results. We describe the different post processing we apply in this section including occlusion filling, median filter and slanted plane smoothing(SPS). And we will compare and analyze the impact of them in the experiments.

\subsection{Disparity Map Occlusion Filling}

We fill the occluded pixels through a simple strategy. First of all, we pick out all the inconsistent pixels using LRC. Rather than correcting them with the smaller value \cite{mozerov2015accurate}, we mark the inconsistent pixels as invalid. Then we fill them with the smallest disparity of the nearest valid pixels.

\subsection{Weighted Median Filter}

As is mentioned in \cite{mozerov2015accurate}, weighted median filter is a robust extension of the bilateral filter and widely used in stereo matching processing. Mozerov et al. \cite{mozerov2015accurate} propose to apply weighted median filter to the whole disparity map which is quite time-consuming. We propose to employ weighted median filtering only to the invalid pixels. After the invalid pixels are filled with occlusion filling, we apply weighted median filter to the disparity map.

\subsection{Slanted Plane Smoothing}
Slanted plane smoothing(SPS) is proposed by Yamaguchi et al. \cite{yamaguchi2014efficient} as a fast approach to dense depth estimate. Their idea is to jointly solve segmentation, occlusion labeling and stereo estimation based on a semi-dense depth map. Though our joint model outputs a dense depth map, we adopt their idea to post process our results. Experiments show that SPS is very effective and efficient.
\section{Experiments and Results}
In this section we evaluate our joint energy minimization approach on the Middlebury stereo benchmark v3. All the evaluations are run and compared under the metric of \textbf{avgErr} with nocc. We start by comparing the locally connected pairwise potential with the fully connected ones. By analyzing the results we show that the joint model outperform every single model, and the LCM and the FCM improves the stereo results of each other.

Then we compare the results of our joint model to the two-step energy minimization approach \cite{mozerov2015accurate} and show that our joint model outperforms theirs. 

Next we analyze and compare the effects of the post-processing techniques we employ. Furthermore, we compare our approach with other approaches on the Middlebury Benchmark v3 in both time and error. 

Our algorithm runs on quarter-size images of the Middlebury dataset and they are up-sampled by the official evaluation toolkit. All the results are obtained on a PC with 3.5GHz Intel i7-4770K CPU and 16GB memory.
 
 In all the experiments we select $\sigma_x = 5$, $\sigma_f = 55$ for Equation (5), $mu_1 = 7$, $mu_2 = 15$, $lambda_1 = 3.5$, $\lambda_2 = 3.0$ and $\lambda_3 = 1.0$ for Equation (8). The parameters are selected to minimize our estimating error of the disparity maps.
 
\subsection{FCM and LCM Analysis}
Here we compare the results of FCM, LCM and FCM+LCM(the joint model, we will also use JEM for abbreviation). The result is shown in Table 1 and Figure 2.
In this section, we only adopt the post-processing strategy of LRC(left-to-right cross checking).

\begin{table}[!hbp]
	    \caption{Compasion of LCM, FCM and JEM(LCM+FCM) models in avgErr. The table shows the error of the results of all 15 data sets and their average. }
	\centering
	\begin{tabular}{lccccccc}
		\toprule
		 Model & LCM & FCM & JEM(FCM+LCM)  \\
		 
		\midrule
		Avg & $7.32$ & $6.58$ & $\mathbf{6.34}$\\
		& & & \\

		Adiron & 2.64 & 2.32 & \textbf{1.93}\\
	
		ArtL & 7.56 & 5.40 & \textbf{4.92}\\
	
		Jadepl & 19.1 & 16.8 & \textbf{15.5}\\
	
		Motor & 4.06 & 3.46 & \textbf{3.26}\\
	
		MotorE & 4.19 & 3.77 & \textbf{3.51}\\
	
		Piano & 5.69 & \textbf{4.52} & 4.62\\
		
		PianoL & 17.1 & \textbf{13.0} & 13.2\\
		
		Pipes & 6.66 & 5.61 & \textbf{5.33}\\
	
		Playrm & 7.99 & \textbf{5.69} & 5.79\\
	
		Playt & \textbf{8.95} & 16.6 & 14.2\\
	
		PlaytP & 3.89 & 4.05 & \textbf{3.36}\\
		
		Recyc & 3.00 & 2.75 & \textbf{2.47}\\
	
		Shelves & 13.7 & \textbf{11.2} & 12.4\\
	
		Teddy & 4.21 & \textbf{3.32} & 4.16\\
		
		Vintage & \textbf{13.3} & 14.2 & 14.9\\
		
		\bottomrule
	\end{tabular}
\end{table}

\begin{figure}[!htbp]
	
	\begin{minipage}[t]{0.25\linewidth}
		\centering
		\includegraphics[width=1.1in]{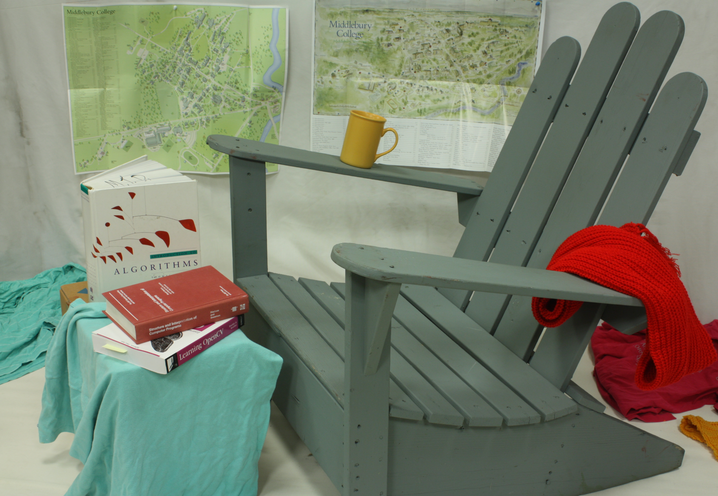}
		\label{fig:side:d11}
	\end{minipage}%
	\begin{minipage}[t]{0.25\linewidth}
		\centering
		\includegraphics[width=1.1in]{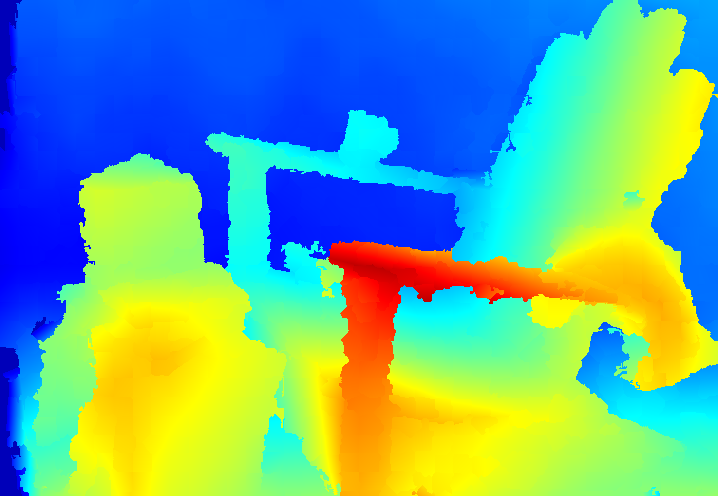}
		\label{fig:side:a11}
	\end{minipage}%
	\begin{minipage}[t]{0.25\linewidth}
		\centering
		\includegraphics[width=1.1in]{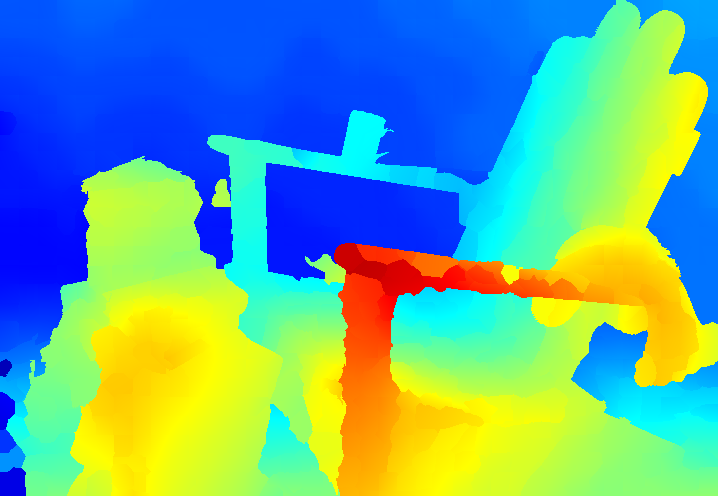}
		\label{fig:side:b11}
	\end{minipage}%
	\begin{minipage}[t]{0.25\linewidth}
		\centering
		\includegraphics[width=1.1in]{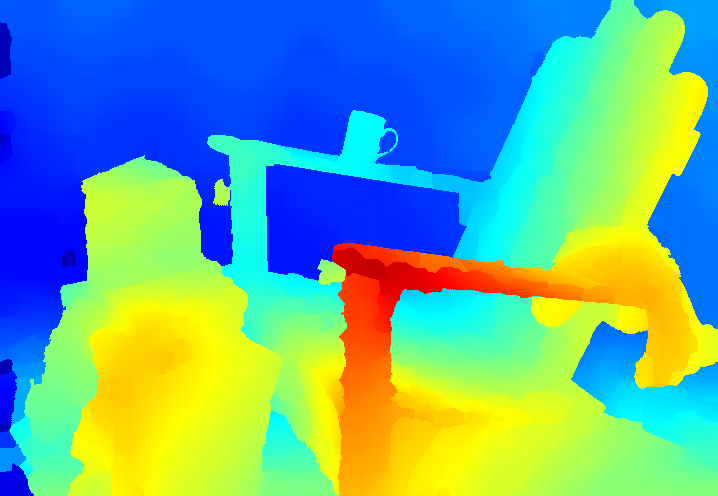}
		\label{fig:side:c11}
	\end{minipage}
	
	\begin{minipage}[t]{0.25\linewidth}
		\centering
		\includegraphics[width=1.1in]{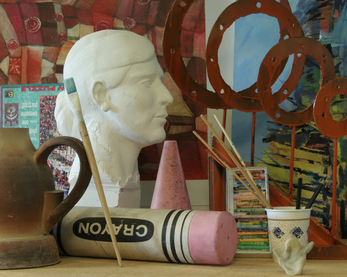}
		\label{fig:side:d21}
	\end{minipage}%
	\begin{minipage}[t]{0.25\linewidth}
		\centering
		\includegraphics[width=1.1in]{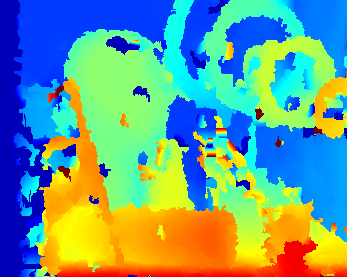}
		\label{fig:side:a21}
	\end{minipage}%
	\begin{minipage}[t]{0.25\linewidth}
		\centering
		\includegraphics[width=1.1in]{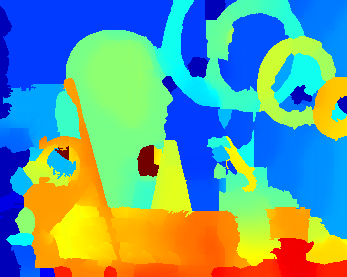}
		\label{fig:side:b21}
	\end{minipage}%
	\begin{minipage}[t]{0.25\linewidth}
		\centering
		\includegraphics[width=1.1in]{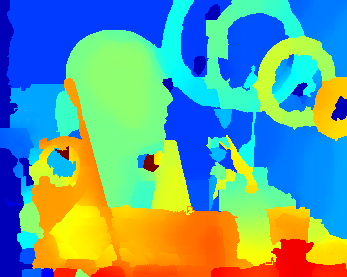}
		\label{fig:side:c21}
	\end{minipage}

\begin{minipage}[t]{0.25\linewidth}
	\centering
	\includegraphics[width=1.1in]{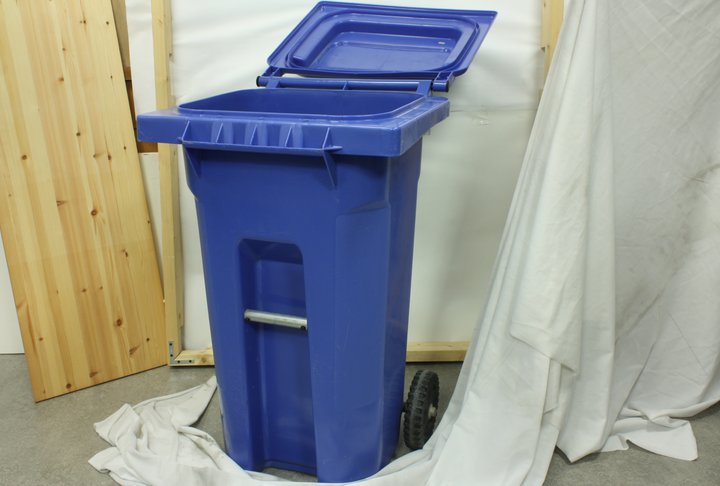}
	\label{fig:side:d31}
	\subcaption{Left}
\end{minipage}%
		\begin{minipage}[t]{0.25\linewidth}
			\centering
			\includegraphics[width=1.1in]{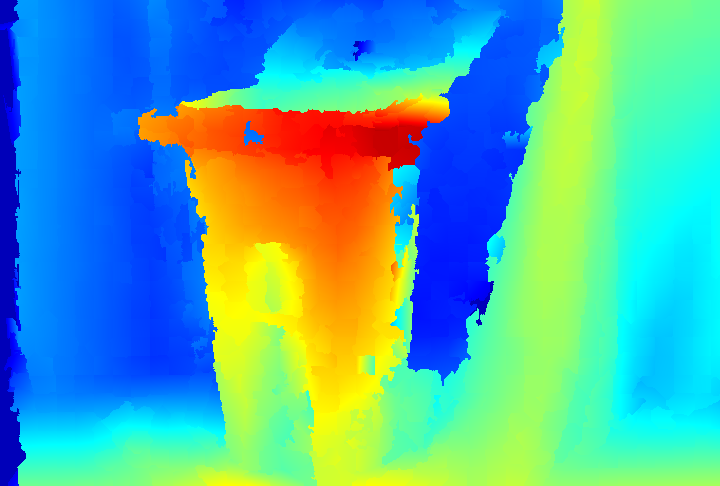}
			\label{fig:side:a31}
			\subcaption{LCM}
		\end{minipage}%
		\begin{minipage}[t]{0.25\linewidth}
			\centering
			\includegraphics[width=1.1in]{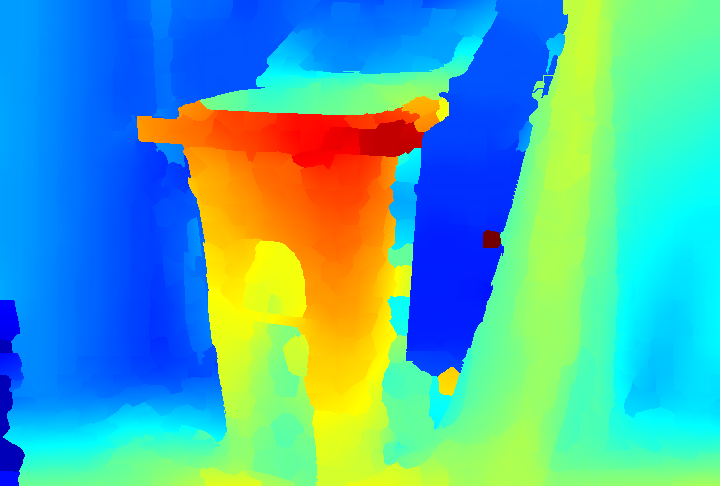}
			\label{fig:side:b31}
			\subcaption{FCM}
		\end{minipage}%
		\begin{minipage}[t]{0.25\linewidth}
			\centering
			\includegraphics[width=1.1in]{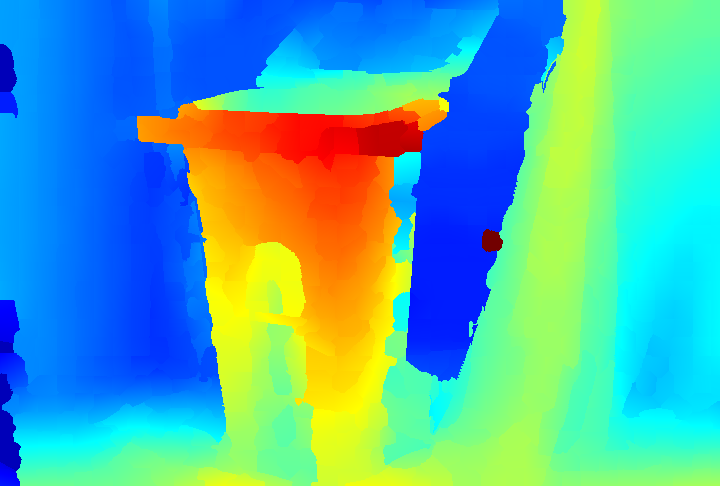}
			\label{fig:side:c31}
			\subcaption{JEM}
		\end{minipage}%
		\caption{Results of LCM, FCM and JEM on Middlebury 3.0 \cite{scharstein2002taxonomy} benchmark (Adirondack, ArtL, Recycle).}
\end{figure}

In average, out joint model outperforms the single FCM or LCM. We can also see from the table that on most data sets, our joint model works better than single FCM or LCM model. Mozerov et al. \cite{mozerov2015accurate} employ the two-step energy minimization approach to improve stereo results in occluded regions. But in our evaluation we only consider non-occuluded regions which is the default of Middlebury benchmark v3, and as a result, we can see that the results of FCM is rather close to JEM compared with LCM. However, in most cases, e.g. Adiron, ArtL, PlaytP, JEM works better than FCM. Moreover, we can see that LCM do achieve better results in small and fine-structured regions while FCM obtains smoother disparity maps (see Figure 2). 

\subsection{Joint Model vs. Two-step Model}

Now we compare out joint model JEM to the previous two-step energy minimization approach(TSGO) in \cite{mozerov2015accurate}. For our JEM model, we only apply one step post-processing: LRC. The TSGO has several steps of post processing. The result is shown in Table 2.
\begin{table}[!hbp]
		\caption{Comparison of TSGO with our joint model JEM in avgErr. Our JEM has only one simple post-processing step: LRC. We can see JEM is better than TSGO in average and especially for data set Jadeplant.}
	\centering
	\begin{tabular}{lcc}
		\toprule
		Algorithm & TSGO & JEM \\
		\midrule
		Avg. & 7.07 & \textbf{6.34} \\
		& & \\
	
		Adiron & 2.02 & \textbf{1.93}\\
	
		ArtL & \textbf{3.07} & 4.92\\
	
		Jadepl & 32.5 & \textbf{15.5}\\
	
		Motor & \textbf{3.12} & 3.26\\
	
		MotorE  & \textbf{2.94} & 3.51\\
	
		Piano & \textbf{2.96} & 4.62\\
	
		PianoL & 16.1 & \textbf{13.2}\\
	
		Pipes &  \textbf{4.90} & 5.33\\

		Playrm &  \textbf{4.02} & 5.79\\
	
		Playt  & 18.7 & \textbf{14.2}\\
	
		PlaytP  & \textbf{2.20} & 3.36\\
	
		Recyc & \textbf{2.33} & 2.47\\
	
		Shelves & \textbf{8.34} & 12.4\\
	
		Teddy  & \textbf{2.46} & 4.16\\
	
		Vintage  & \textbf{12.6} & 14.9\\
		\bottomrule
	\end{tabular}
\end{table}

Even with only a single simple step of post-processing, our model outperforms TSGO in average. The result shows that the joint model has advantages over the splitted two-step model. Especially for the data set Jadeplant, TSGO has average error 32.5 while our JEM achieves 15.5 which is surprisingly only half of it. 

After we add more post-processing steps, our results are further improved(see Table 4).
\subsection{Post-processing Analysis}

In this section, we compare the results of different post-processing techniques. First we utilize left-to-right cross checking(LRC), and then we apply slanted plane smoothing(SPS). In the end we employ occlusion filling(OF) with weighted median filter(WMF) in addition to the previous SPS. The result is shown in Table 3.
\begin{table}[!hbp]	
		\caption{Comparison of different post-processing techniques in avgErr. LRC refers to left-to-right cross checking. OF refers to occlusion filling. WMF refers to weighted median filter and SPS mean slanted plane smoothing.}
	\centering
	\begin{tabular}{lccc}
		\toprule
		Algorithm & LRC & SPS & WMF + OF + SPS \\
		\midrule
		Avg & 6.34 & 5.52 & \textbf{5.33} \\
			& & & 
		\\ 
	
		Adiron & \textbf{1.93} & 1.96 & 1.96\\
	
		ArtL & 4.92 & 4.90 & \textbf{4.32}\\
		
		Jadepl & 15.5 & 16.4 & \textbf{15.5}\\
	
		Motor & 3.26 & 3.28 & \textbf{3.23}\\
	
		MotorE & 3.51 & \textbf{3.25} & 3.26\\
		
		Piano & 4.62 & 3.60 & \textbf{3.57}\\
		
		PianoL & 13.2 & 12.4 & \textbf{12.1}\\
	
		Pipes & 5.33 & 5.33 & \textbf{5.14}\\
	
		Playrm &  5.79 & 5.69 & \textbf{5.69}\\
	
		Playt & 14.2 & 6.73 & \textbf{6.71}\\
		
		PlaytP & 3.36 & 2.62 & \textbf{2.56}\\
	
		Recyc & 2.47 & 2.40 & \textbf{2.34}\\
	
		Shelves & 12.4 & 9.49 & \textbf{9.28}\\
		
		Teddy &  4.16 & 3.39 & \textbf{3.17}\\
	
		Vintage & 14.9 & 9.40 & \textbf{9.40}\\
		
		\bottomrule
		
	\end{tabular}
\end{table}

From Table 3 we can see that in average WMF + OF + SPS effects the best. This is of no surprise because more post processing usually lead to better results.  Surprisingly we see little improvement on several data sets when adding WMF and OF to SPS, and even worse result on MotorE. We observe that the Middlebury stereo dataset mainly contains images of indoor scenes where walls and planar objects occupy the majority of the scenes. Thus stereo methods which rely on planar elements have natural advantages over the others. As a result, the SPS can play a key role in the results and it improves the original stereo results a lot.

\subsection{More Comparisons}

In this section, we compare our approach to the other approaches on the table of Middlebury benchmark v3. We compare our results with the others in both aspects of running time and accuracy.

\begin{figure}[!htbp]

	\begin{minipage}[t]{0.25\linewidth}
		\centering
		\includegraphics[width=1.1in]{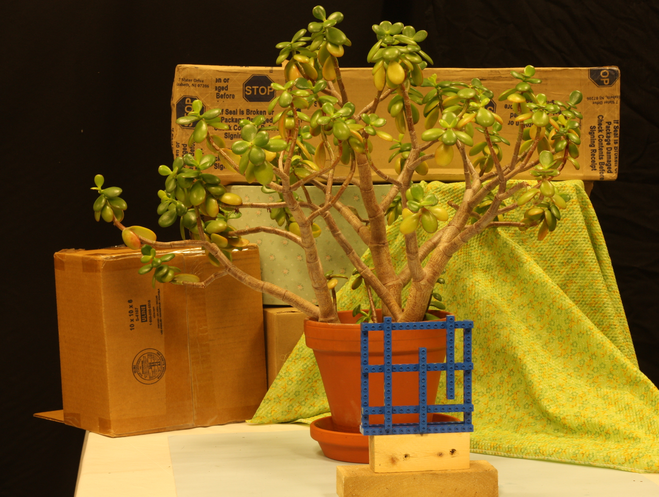}
		\label{fig:side:a1}
	\end{minipage}%
	\begin{minipage}[t]{0.25\linewidth}
		\centering
		\includegraphics[width=1.1in]{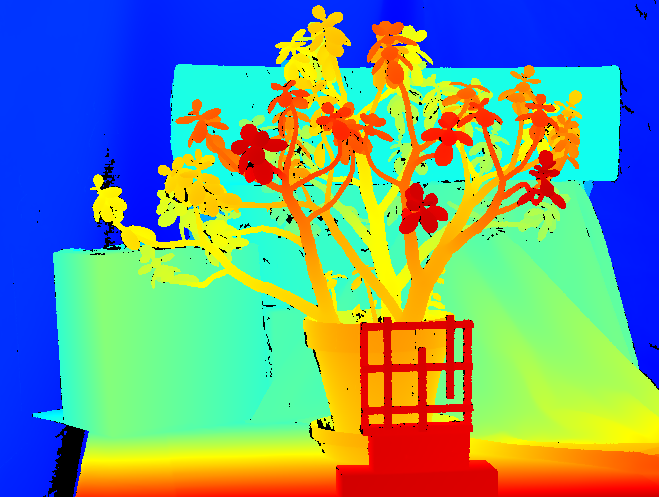}
		\label{fig:side:b1}
	\end{minipage}%
	\begin{minipage}[t]{0.25\linewidth}
		\centering
		\includegraphics[width=1.1in]{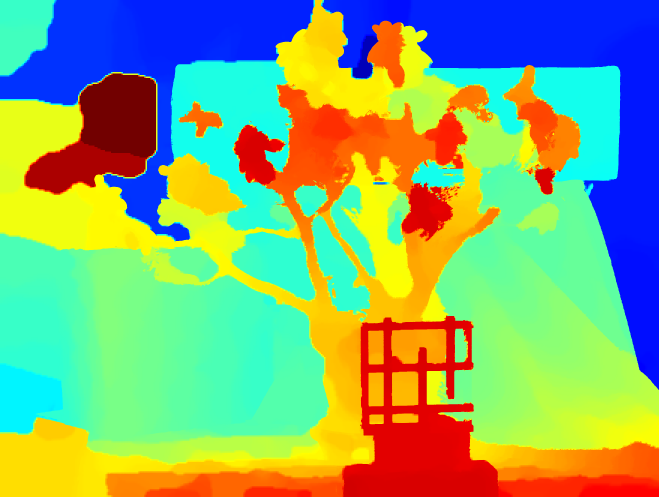}
		\label{fig:side:c1}
	\end{minipage}%
	\begin{minipage}[t]{0.25\linewidth}
		\centering
		\includegraphics[width=1.1in]{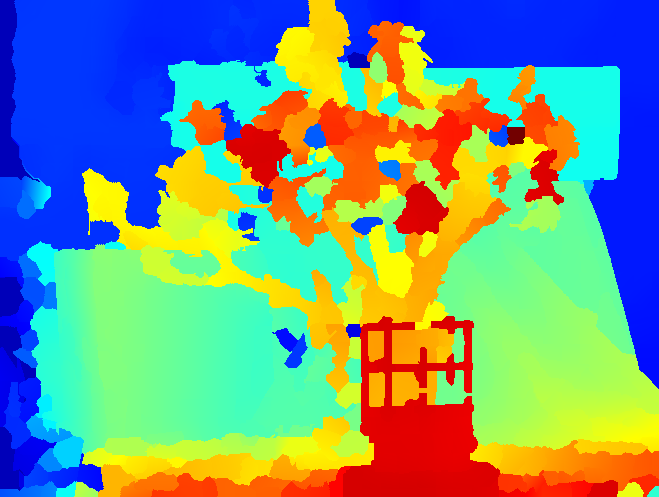}
		\label{fig:side:d1}
	\end{minipage}
	
	\vspace{1pt}
	
		\begin{minipage}[t]{0.25\linewidth}
			\centering
			\includegraphics[width=1.1in]{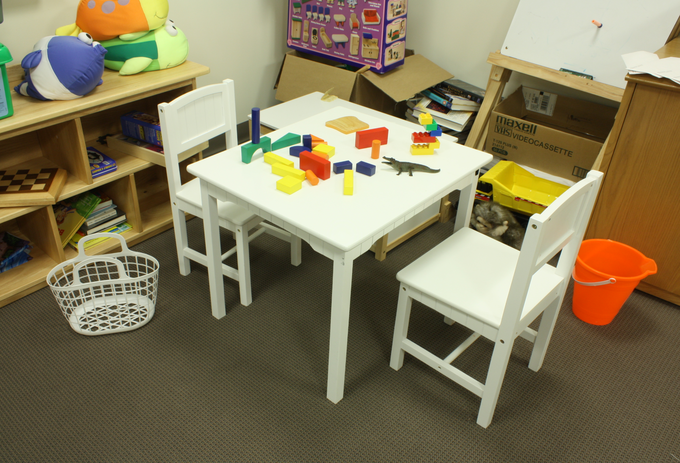}
			\label{fig:side:a2}
		\end{minipage}%
		\begin{minipage}[t]{0.25\linewidth}
			\centering
			\includegraphics[width=1.1in]{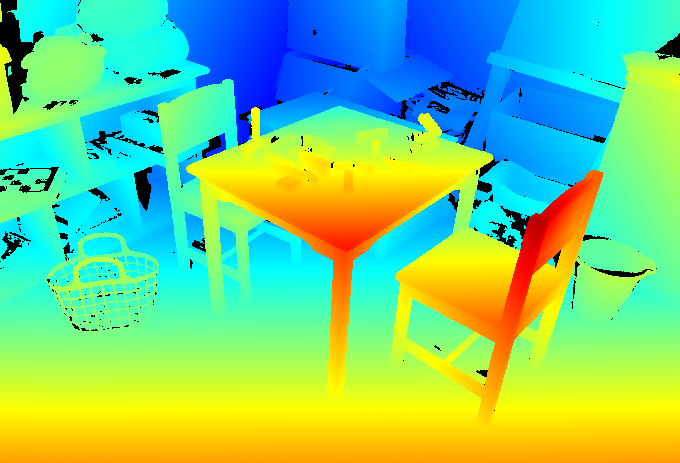}
			\label{fig:side:b2}
		\end{minipage}%
		\begin{minipage}[t]{0.25\linewidth}
			\centering
			\includegraphics[width=1.1in]{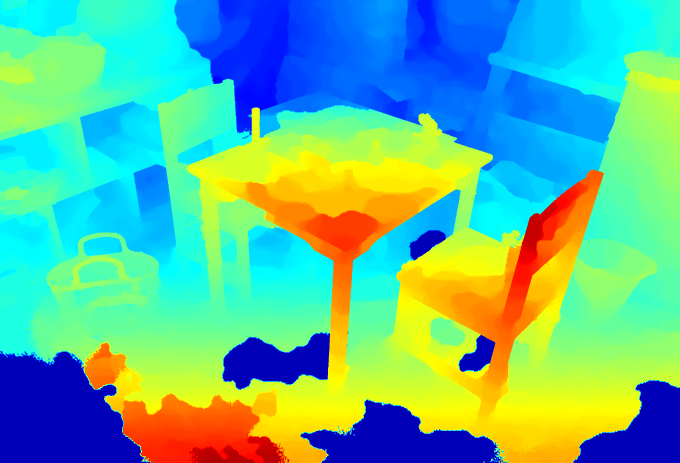}
			\label{fig:side:c2}
		\end{minipage}%
		\begin{minipage}[t]{0.25\linewidth}
			\centering
			\includegraphics[width=1.1in]{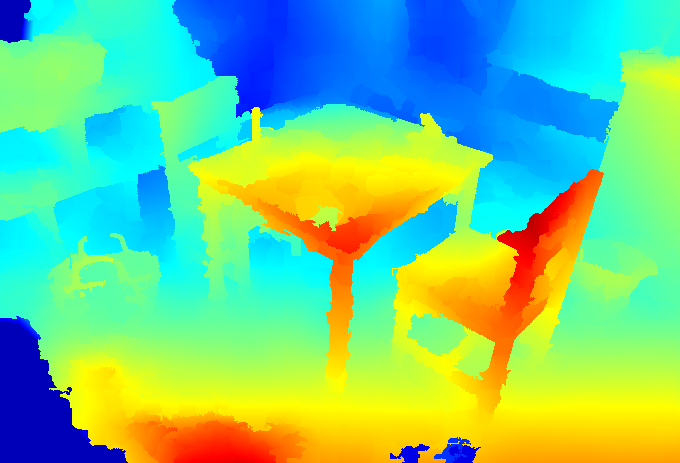}
			\label{fig:side:d2}
		\end{minipage}
		
			\begin{minipage}[t]{0.25\linewidth}
				\centering
				\includegraphics[width=1.1in]{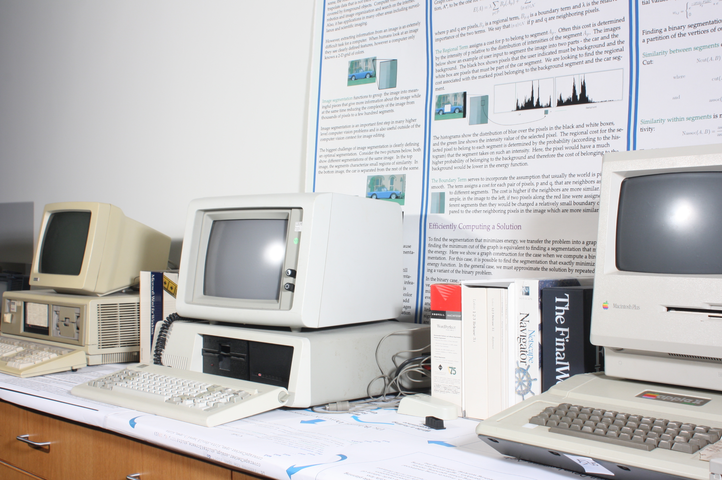}
				\label{fig:side:a}
				\subcaption{Left}
			\end{minipage}%
			\begin{minipage}[t]{0.25\linewidth}
				\centering
				\includegraphics[width=1.1in]{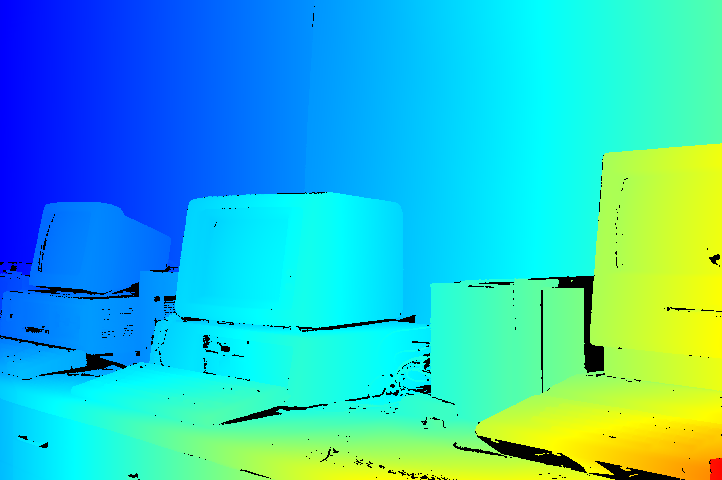}
				\label{fig:side:b}
				\subcaption{GT}
			\end{minipage}%
			\begin{minipage}[t]{0.25\linewidth}
				\centering
				\includegraphics[width=1.1in]{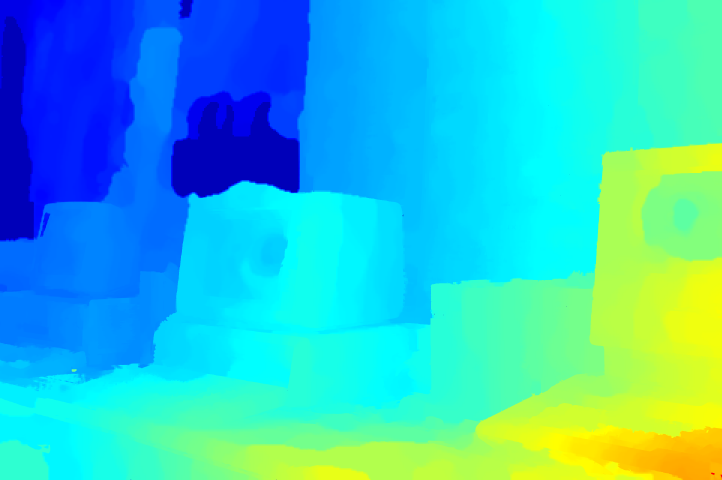}
				\label{fig:side:c}
				\subcaption{TSGO}
			\end{minipage}%
			\begin{minipage}[t]{0.25\linewidth}
				\centering
				\includegraphics[width=1.1in]{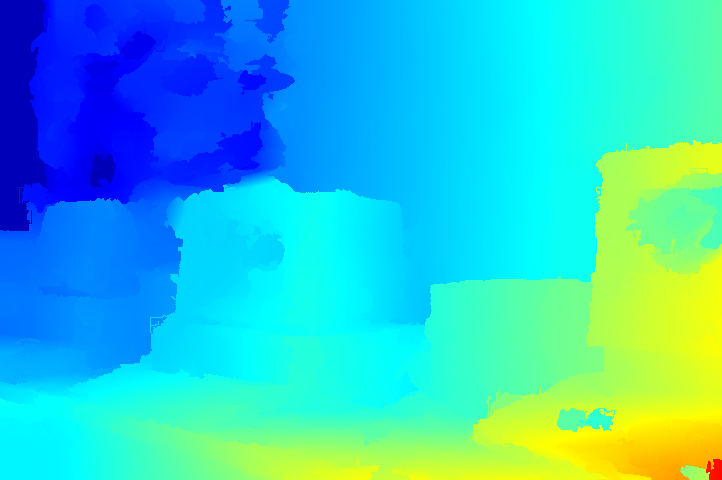}
				\label{fig:side:d}
				\subcaption{JEM(Ours)}
			\end{minipage}
			\caption{Results on the Middlebury 3.0 \cite{scharstein2002taxonomy} benchmark with ground truth (Jadeplant, Playtable, Vintage).}
\end{figure}
\subsubsection{Error}
We compare our results with several other methods(most of them are recently proposed).
\begin{table}[!hbp]	
		\caption{Comparison with other approaches in avgErr. Notice first that JEM outperforms TSGO especially on Jadeplant, Playtable and Vintage (see Figure 3). Our result are also better than that of MC-CNN-art on Jadeplant and Vintage.}
	\centering
	\begin{tabular}{lcccp{0.6cm}ccc}
		\toprule
		Alg. & \scriptsize{MC-CNN-art \cite{vzbontar2015stereo}} & \scriptsize{MeshStereo \cite{zhang2015meshstereo}} & JEM & \scriptsize{SGM \cite{hirschmuller2005accurate}} & \small{TSGO} & \scriptsize{LPS \cite{sinha2014efficient}}\\
	
	\midrule
		Avg  & 3.82 & 4.38 & 5.33 & 5.82 & 7.07 & 7.58 \\
		& & & & & & 
		\\ 
	
		Adiron & 0.76 & 1.55 & 1.96 & 4.62 & 2.02 & 2.15 \\ 
		
		ArtL & 2.55 & 2.93 & 4.32 & 2.34 & 3.07 & 12.6 \\
		
		Jadepl & 16.3 & 16.2 & 15.5 & 13.2 & 32.5 & 15.4 \\
		
		Motor & 1.27 & 2.83 & 3.23 & 2.97 & 3.12 & 1.45\\
	
		MotorE & 1.27 & 3.42 & 3.26 & 2.49 & 2.94 & 27.6 \\
	
		Piano & 1.83 & 2.71 & 3.57 & 3.24　& 2.96 & 3.40 \\
	
		PianoL & 5.07 & 4.94 & 12.1 & 5.18 & 16.1 & 20.1 \\

		Pipes & 2.29 & 6.03 & 5.14 & 5.35 & 4.90 & 3.63 \\
	
		Playrm & 2.27 & 4.57 & 5.69 & 4.40 & 4.02 & 4.60 \\
	
		Playt & 3.11 & 3.82 & 6.71 & 23.9 & 18.7 & 3.24 \\
	
		PlaytP & 3.03 & 2.02 & 2.56 & 3.61 & 2.20 & 3.15 \\
	
		Recyc & 2.48 & 1.65 & 2.34 & 4.08 & 2.33 & 2.44 \\
	
		Shelves & 4.41 & 10.7 & 9.28 & 8.83 & 8.34 & 9.51 \\
	
		Teddy &  1.12 & 1.09 & 3.17 & 1.52 & 2.46 & 1.83\\
		
		Vintage & 14.8 & 4.64 & 9.40 & 16.2 & 12.6 & 4.76\\
		
		\bottomrule
		
	\end{tabular}
\end{table}
	Table 4 lists the results of several approaches on the Middlebury benchmark v3. First of all, we can see our result outperform TSGO on many data sets and especially on Jadeplant, Playtable and Vintage. On the other data sets the results are rather close. We also observe that different approaches have advantages on different data sets. Even though LPS has the worst average result, it shows the best effect on Vintage. Our JEM also outperforms MC-CNN-art on Jadeplant and Vintage, and outperforms MeshStereo on several data sets including Jadeplant and Shelves.
	
\subsubsection{Runtime Analysis}

\begin{table}[!hbp]	
		\caption{Comparison with other approaches in time(in seconds). We can see that the CNN based method is extremely slow. MeshStereo and TSGO are also slower than JEM. Only LPS is faster than JEM.}
	\centering
	\begin{tabular}{lcccccc}
		\toprule
		Alg. & MC-CNN-art & MeshStereo & JEM & SGM & TSGO & LPS\\
		\midrule
		Runtime & 106 & 62.0 & 20.8 & 52.3 & 53.8 & 9.35 \\
		\bottomrule
	\end{tabular}
\end{table}

Comparing our approach with others in running time (see Table 5), we can see our approach is efficient compared with high-accuracy methods(MC-CNN-art and MeshStereo). JEM is also efficient when compared with approaches that have close error(SGM, TSGO). Though it takes a lot more time than LPS, it is more accurate than LPS. 
\section{Conclusion}

We propose a joint energy minimization approach(JEM) based on the two-step energy minimization method(TSGO) \cite{mozerov2015accurate}. JEM is inspired by the observations that FCM is more suitable for large and flat structures while LCM achieves better results on small and fine-structure regions. 

The joint energy term consists of two parts: a global pairwise potential and a local one. JEM enables the effect of both models simultaneously and shows superior results to TSGO. In order to solve the joint model efficiently, we employ mean field approximation \cite{krahenbuhl2012efficient} to inference the posterior probability of our model. The mean field inference to the joint model is much more efficient than separate solutions to each models. As a result, JEM is faster and more effective than TSGO which is shown by experiments on the Middlebury benchmark v3. JEM is also comparable to several the state-of-the-art approaches in both computational time and estimation error.

\bibliographystyle{plain}
\bibliography{jem.bib}
\end{document}